\documentclass[11pt,a4paper]{article}
\usepackage[hyperref]{acl2019}
\usepackage{times}
\usepackage{latexsym}
\usepackage{graphicx}
\usepackage{url}
\usepackage{algorithm}
\usepackage{algpseudocode}
\usepackage{mathrsfs}
\usepackage{amsmath}
\usepackage{multirow}
\usepackage{amssymb}
\usepackage{enumitem}
\aclfinalcopy 

   \makeatletter
\def\@fnsymbol#1{\ensuremath{\ifcase#1\or \dagger\or \ddagger\or
   \mathsection\or \mathparagraph\or \|\or **\or \dagger\dagger
   \or \ddagger\ddagger \else\@ctrerr\fi}}
    \makeatother

\title{Personalizing Dialogue Agents via Meta-Learning}

\author{Zhaojiang Lin\thanks{$^\dagger$ These two authors contributed equally.}, Andrea Madotto$^\dagger$, Chien-Sheng Wu, Pascale Fung\\
  Center for Artificial Intelligence Research (CAiRE) \\
  Department of Electronic and Computer Engineering \\
  The Hong Kong University of Science and Technology, Clear Water Bay, Hong Kong \\
  {\tt [zlinao,amadotto,cwuak,pascale]@ust.hk} }
  
\date{}

\begin{document}
\maketitle
\begin{abstract}


Existing personalized dialogue models use human designed persona descriptions to improve dialogue consistency. Collecting such descriptions from existing dialogues is expensive and requires hand-crafted feature designs. In this paper, we propose to extend Model-Agnostic Meta-Learning (MAML)~\cite{finn2017model} to personalized dialogue learning without using any persona descriptions. Our model learns to quickly adapt to new personas by leveraging only a few dialogue samples collected from the same user, which is fundamentally different from conditioning the response on the persona descriptions. Empirical results on Persona-chat dataset~\cite{personachat} indicate that our solution outperforms non-meta-learning baselines using automatic evaluation metrics, and in terms of human-evaluated fluency and consistency.


\end{abstract}

\section{Introduction}
There is a growing interest in learning personalized chit-chat dialogue agents for making chat-bots more consistent. Recently, a multi-turn conversational dataset called Persona-chat~\cite{personachat} has been released, where two speakers are paired and a persona description (4-5  sentences) is randomly assigned to each of them. For example, ``\textit{I am an old man}'' and ``\textit{I like to play football}'' are one of the possible persona descriptions provided to the speaker. By conditioning the response generation on the persona descriptions, a chit-chat model is able to produce a more persona consistent dialogue~\cite{personachat}. 

However, it is difficult to capture a persona just by using few sentences, and collecting a non-synthetic set of persona descriptions from a real human-human conversation, e.g., Reddit, is challenging as well since it requires hand-crafted feature designs~\cite{millionspersona}. 
In light of this, we propose to leverage a set of dialogues done by the same persona directly, instead of using its persona descriptions, to generate a more consistent response.

\begin{figure}[t]
    \centering
    \includegraphics[width=\linewidth]{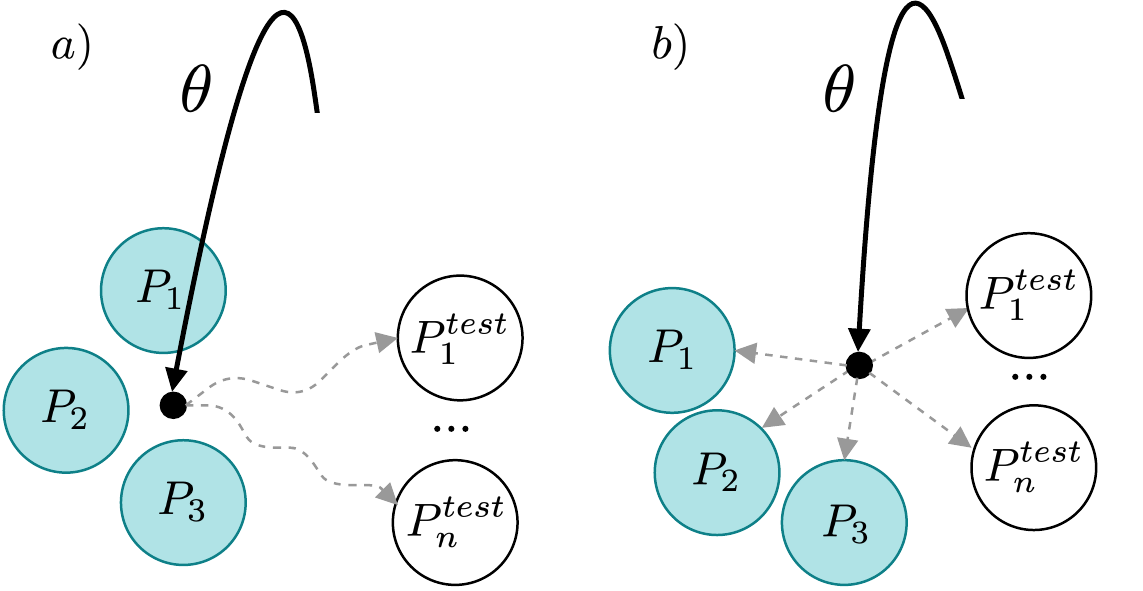}
    \caption{The difference between finetuning from a) joint training on all personas and b) meta-learning persona. The solid line represents the optimization path of the initial parameters and dashed line the fine-tuning path. Meta-learned initial parameters can faster adapt to a new persona.}
    \label{fig:maml}
\end{figure}

We consider learning different personas as different tasks via meta-learning algorithms, which is fundamentally different from optimizing the model to represent all the personas. 
A high-level intuition of the difference between these two approaches is shown in Figure~\ref{fig:maml}.
We aim to learn a persona-independent model that is able to quickly adapt to a new persona given the dialogues.
We formulate this task as a few-shot learning problem, where $K$ dialogues are used for training and the remaining for the test.
Hence, we expect to learn initial parameters of a dialogue model that can quickly adapt to the response style of a certain persona just by using few dialogues. 

The main contribution of this paper is to cast the personalized dialogue learning as a meta-learning problem, which allows our model to generate personalized responses by efficiently leveraging only a few dialogue samples instead of human-designed persona descriptions. Empirical results show that our solution outperforms joint training, in terms of human-evaluated fluency and consistency.

\section{Personalized Dialogue Learning}
\subsection{Persona-conditioned dialogue}
In Persona-chat dataset~\cite{personachat}, a dialogue is defined as a set of utterances $U=\{u_1, \dots, u_n\}$ and a persona description is defined as a set of sentences $P=\{p_1,\dots,p_m\}$. A personalized dialogue model $f_{\theta}$ is trained to produce a response $Y=u_t$ conditioned on previous utterances $X=\{u_1, \dots, u_{t-1}\}$ and persona sentences $P$:  
\begin{equation}
f_{\theta}(Y | X,P ; \theta)= p\left( u_t | u_{1 : t-1}, p_{1 :m} ; \theta\right)
\label{dialog_persona}
\end{equation}

\subsection{Persona-agnostic dialogue}
Instead of conditioning our response on the persona sentences, we first adapt $\theta$ to the set of dialogue made by a persona $P$ and then we only use the dialogue history to condition our response. Eq. (\ref{dialog_persona}) becomes:
\begin{equation}
f_{\theta}(Y | X ; \theta)= p\left( u_t | u_{1 : t-1} ; \theta\right)
\label{dialog}
\end{equation}
Therefore, we define the set of dialogues of a persona $P$ as $\mathcal{D}_p= \{U_1, \dots, U_k \}$. Conceptually, a model $f_{\theta}$ is expected to generate personalized response after being trained with a few dialogues example from $\mathcal{D}_p$. 
The main idea of our work is to use Model-Agnostic Meta-Learning (MAML)~\cite{finn2017model} to learn an initial set of parameters that can quickly learn a persona from few dialogues sample. 
We refer to the proposed meta-learning method for persona dialogues as Persona-Agnostic Meta-Learning (PAML).

\paragraph{Persona-agnostic meta-learning (PAML)} 
We define the persona meta-dataset as $\mathscr{D} = \{\mathcal{D}_{p_1}, \dots, \mathcal{D}_{p_z}\}$, where $z$ is the number of persona. Before training, $\mathscr{D}$ is split into $\mathscr{D}_{train}, \mathscr{D}_{valid}, \mathscr{D}_{test}$.  
\begin{algorithm}[t]
{\selectfont
\caption{Persona-Agnostic Meta-Learning}
\label{alg1}
\textbf{Require:} $\mathscr{D}_{train}$\\
\textbf{Require:} $\alpha, \beta$: step size hyperparameters
\begin{algorithmic}[1]
\State Randomly initialize $\theta$
\While{not done}
  \State Sample batch of persona $\mathcal{D}_{p_i} \sim \mathscr{D}_{train}$
  \For{\textbf{all} $\mathcal{D}_{p_i}$}
    \State $( \mathcal{D}^{train}_{p_i}, \mathcal{D}^{valid}_{p_i})\sim$ $ \mathcal{D}_{p_i}$
      \State Evaluate $\nabla_{\theta}\mathcal{L}_{\mathcal{D}^{train}_{p_i}}(f_{\theta})$ using $\mathcal{D}^{train}_{p_i}$
      \State Compute adapted parameters with \phantom . \phantom .\phantom .\phantom .\phantom .\phantom .\phantom .\phantom .\phantom .\phantom .\phantom .\phantom .\phantom .gradient descent:
       \phantom . \phantom .\phantom .\phantom .\phantom .\phantom .\phantom .\phantom .\phantom .\phantom .\phantom .\phantom .\phantom .
       \phantom . \phantom .\phantom .\phantom .\phantom .\phantom .\phantom .\phantom .\phantom .\phantom .\phantom .\phantom .\phantom .
       \phantom .\phantom .\phantom .\phantom .\phantom .\phantom .\phantom .\phantom .\phantom .\phantom .\phantom .
       $\theta _ { p_i } ^ { \prime } = \theta - \alpha \nabla _ { \theta } \mathcal { L } _ { \mathcal { D }^{train} _ { p_i } } \left( f _ { \theta } \right)$
  \EndFor
  \State $\theta \leftarrow \theta-\beta\sum_{\mathcal{D}_{p_i}\sim \mathscr{D}_{train}} \nabla_{\theta}\mathcal{L}_{\mathcal{D}^{valid}_{p_i}}\left(f_{\theta_{p_i}^{\prime}}\right)$  
\EndWhile
\end{algorithmic}
}
\end{algorithm}
\begin{table*}[t]
\centering
\begin{tabular}{r|ccc|cc}
\hline
\multicolumn{1}{l|}{} & \multicolumn{3}{c|}{\textit{\textbf{Automatic}}} & \multicolumn{2}{c}{\textit{\textbf{Human}}} \\ \hline
\multicolumn{1}{l|}{} & \textbf{PPL} & \textbf{BLEU} & \textit{\textbf{C}} & \textbf{Fluency} & \textbf{Consistency} \\ \hline
\textit{Human} & - & - & 0.33 & 3.434 & 0.234 \\ 
\textit{Dialogue+Persona} & \textbf{30.42} & \textbf{1.00} & 0.07 & 3.053 & 0.011 \\ \hline
\textit{Dialogue} & 36.75 & 0.64 & -0.03 & - & - \\ 
\textit{Dialogue+Fine-tuning} & 32.96 & 0.90 & 0.00 & 3.103 & 0.038 \\
\textit{PAML} & 41.64 & 0.74 & \textbf{0.20} & \textbf{3.185} &\textbf{ 0.197} \\ \hline
\end{tabular}
\caption{Results of automatic and human evaluation: \textit{\textbf{PAML}} vs \textit{\textbf{Dialogue+Persona}} shows the our approach can achieve good consistency by using few dialogues instead of conditioning on the persona description, \textit{\textbf{PAML}} vs \textit{\textbf{Dialogue+Fine-tuning}} shows the effectiveness of meta-learning approach in personalizing dialogue model.}
\label{tab:res}
\end{table*}
For each training epoch, we uniformly sample a batch of personas $\mathcal{D}_{p_i}$ from $\mathscr{D}_{train}$, then from each persona in $\mathcal{D}_{p_i}$ we sample a set of dialogues as training $\mathcal{D}^{train}_{p_i}$, and another set of dialogues as validation $\mathcal{D}^{valid}_{p_i}$.
After $t$ iterations of training on $\mathcal{D}^{train}_{p_i}$, the dialogue model $f_{\theta}$, parameterized by $\theta$, is updated to $\theta_{p_{i}}^{\prime}$ by standard gradient descent,
\begin{equation}
    \theta_{p_i}^{\prime}=\theta-\alpha \nabla_{\theta} \mathcal{L}_{\mathcal{D}^{train}_{p_i}}\left(f_{\theta}\right)
\end{equation}
where $\alpha$ is learning of the inner optimization, and $\mathcal{L}_{\mathcal{D}^{train}_{p_i}}$ the training loss. Specifically, cross-entropy loss is used for training the response generation:
\begin{equation}
    \mathcal{L}_{\mathcal{D}_{p_i} }\left(f_{\theta}\right) = -\sum_{\mathcal{D}_{p_i}}\log p\left( u_t | u_{1 : t-1} ; \theta\right)
\end{equation}
The meta-learning model is then trained to maximize the performance of the adapted model $f_{\theta_{p_i}^{\prime}}$ to the unseen dialogues in  $\mathcal{D}^{valid}_{p_i}$. Following \citet{finn2017model}, we define the meta-objective as:
\begin{align}
    \min _{\theta}&\sum_{\mathcal{D}_{p_i}\sim \mathscr{D}_{train}}\mathcal{L}_{\mathcal{D}^{valid}_{p_i}}\left(f_{\theta_{p_i}^{\prime}}\right) = \nonumber \\ &\sum_{\mathcal{D}_{p_i}\sim \mathscr{D}_{train}}\mathcal{L}_{\mathcal{D}^{valid}_{p_i}}\left(f_{{\theta-\alpha} \nabla_\theta \mathcal{L}_{\mathcal{D}^{train}_{pi}}\left(f_{\theta}\right)}\right) \label{EQ_maml}
\end{align}
where $\mathcal{L}_{\mathcal{D}^{valid}_{p_i}}\left(f_{\theta_{p_i}^{\prime}}\right)$ is the loss evaluated on $\mathcal{D}^{valid}_{p_i}$. 
For optimizing Eq.(\ref{EQ_maml}), we apply again stochastic gradient descent on the meta-model parameters $\theta$ by computing the gradient of $\mathcal{L}_{\mathcal{D}^{valid}_{p_i}}\left(f_{\theta_{p_i}^{\prime}}\right)$, which is:
\begin{equation}
    \theta \leftarrow \theta-\beta\sum_{\mathcal{D}_{p_i}\sim \mathscr{D}_{train}} \nabla_{\theta}\mathcal{L}_{\mathcal{D}^{valid}_{p_i}}\left(f_{\theta_{p_i}^{\prime}}\right)
\end{equation}
where $\beta$ is meta-learning rate. This process requires second order optimization partial derivatives, which can be computed by any automatic differentiation library (e.g. PyTorch, Tensorflow etc.). A summary of the training procedure is shown in Algorithm \ref{alg1}.

\section{Experiment and Results}
The experiments are conducted using Persona-chat~\cite{personachat}. 
To create the meta-sets $\mathscr{D}$, we match the dialogues by their persona description separately for train, validation and test, by following the same persona split as in ~\citet{personachat}. On average each persona description has 8.3 unique dialogues. In the Appendix, we report the number of dialogue distribution. 

\paragraph{Experimental setting} In our experiments, we compared different training settings:
(\textit{\textbf{Dialogue}}) a model trained using dialogue history, as in Eq.(\ref{dialog}); 
(\textit{\textbf{PAML}}) a meta-trained model as in Eq.(\ref{EQ_maml}), where we test each set $\mathcal{D}_{p_i}\in \mathscr{D}_{test}$ by selecting one dialogue and training with all the others. To elaborate, suppose we are testing $U_t \in \mathcal{D}_{p_i}$ then we first fine-tuning using all the dialogues in $\mathcal{D}_{p_i}\setminus U_t$, and then test on $U_t$. This process is repeated for all the dialogues in $ \mathcal{D}_{p_i}$. 
(\textit{\textbf{Dialogue+Fine-tuning}}) we use the same testing as \textit{PAML} but on a model trained as \textit{Dialogue}.
We also report a trained model that assumes persona description is available and we refer it as (\textit{\textbf{Dialogue+Persona}}).\\ 


\paragraph{Implementation details} We implemented $f_{\theta}$ using a standard Transformer architecture~\cite{transformer} with pre-trained Glove embedding~\cite{pennington2014glove}~\footnote{The model and the pre-processing scripts are available at \url{https://github.com/HLTCHKUST/PAML}}. For the standard training, we used Adam~\cite{kingma2014adam} optimizer with a warm-up learning rate strategy, and a batch size of 32. Instead, in meta-training, we used SGD for the inner loop and Adam for the outer loop with learning rate $\alpha=0.01$ and $\beta=0.0003$ respectively, and batch size of 16 for both. In all the model we used beam search with beam size 5.

\subsection{Evaluation metric}
The objective of the evaluation is to verify whether PAML can produce a more consistent response with reference to the given dialogue and persona description (even though is not seen). To do so, we employ both automatic and human evaluation. 
\paragraph{Automatic} We report perplexity and BLEU score~\cite{papineni2002bleu} of the generate sentences against the human-generated prediction. 
Aside of standards evaluation metrics, we also train a Natural Language Inference (NLI) model using Dialog NLI~\cite{dnli} dataset, a recently proposed corpus based on Persona dataset, with NLI annotation between persona description sentences and dialogues utterance. We fine-tune a pre-trained BERT model~\cite{devlin2018bert} using the DNLI corpus and achieve a test set accuracy of 88.43\%, which is aligned to the best-reported model ESIM~\cite{P17-1152} in \citet{dnli} (with 88.20\% accuracy). Then, we defined a new evaluation metric for dialogue consistency as follow:
\begin{align}
    &\text{\textbf{NLI}}(u, p_j) = \Bigg\{
  \begin{array}{rcr}
    1 & \text{if $u$ entails $p_j$} \\
    0 & \text{if $u$ is independent to $p_j$} \\
    -1 & \text{if $u$ contradicts $p_j$} \\
  \end{array} \nonumber
  \\
  &\text{\textit{\textbf{C}}}(u) = \sum_j^{m} \text{\textbf{NLI}}(u, p_j)
\end{align}
where $u$ is a generated utterance and the $p_j$ is one sentence in the persona description. Hence, having a higher consistency \textit{\textbf{C}} score means having a more persona consistent dialogue response. 

\paragraph{Human} Since automatic evaluation performs poorly in this task~\cite{D16-1230}, we perform a human evaluation using crowd-sourced workers. We randomly selected 300 generated response examples from 10 unique personas and we asked each worker to evaluate fluency (1 to 5) and consistency of the generated response with respect to the dialogue history and the respective persona description. We asked the workers to assign a score of 1, 0 or -1 for consistent, neutral, and contradicts respectively, the full instruction set is available in the Appendix.

\subsection{Results} 
Table~\ref{tab:res} shows both automatic and human evaluation results. \textit{PAML} achieve consistently better results in term of dialogue consistency in both automatic and human evaluation. The latter also shows that all the experimental settings have comparable fluency scores, where instead perplexity and BLEU score are lower in \textit{PAML}. This confirms that these measures are not correlated to human judgment~\cite{D16-1230}. For completeness, we also show generated responses examples from \textit{PAML} and baseline models in Appendix. 

On the other hand, the human evaluated consistency is aligned to the \textit{\textbf{C}} score, which confirms the meaningfulness of the defined measure. This agrees with results of \citet{dnli}, where the authors showed that by re-ranking the beam search hypothesis using the DNLI score (i.e. \textit{\textbf{C}} score), they achieved a substantial improvement in dialogue consistency. 

\begin{figure}
    \centering
    \includegraphics[width=\linewidth]{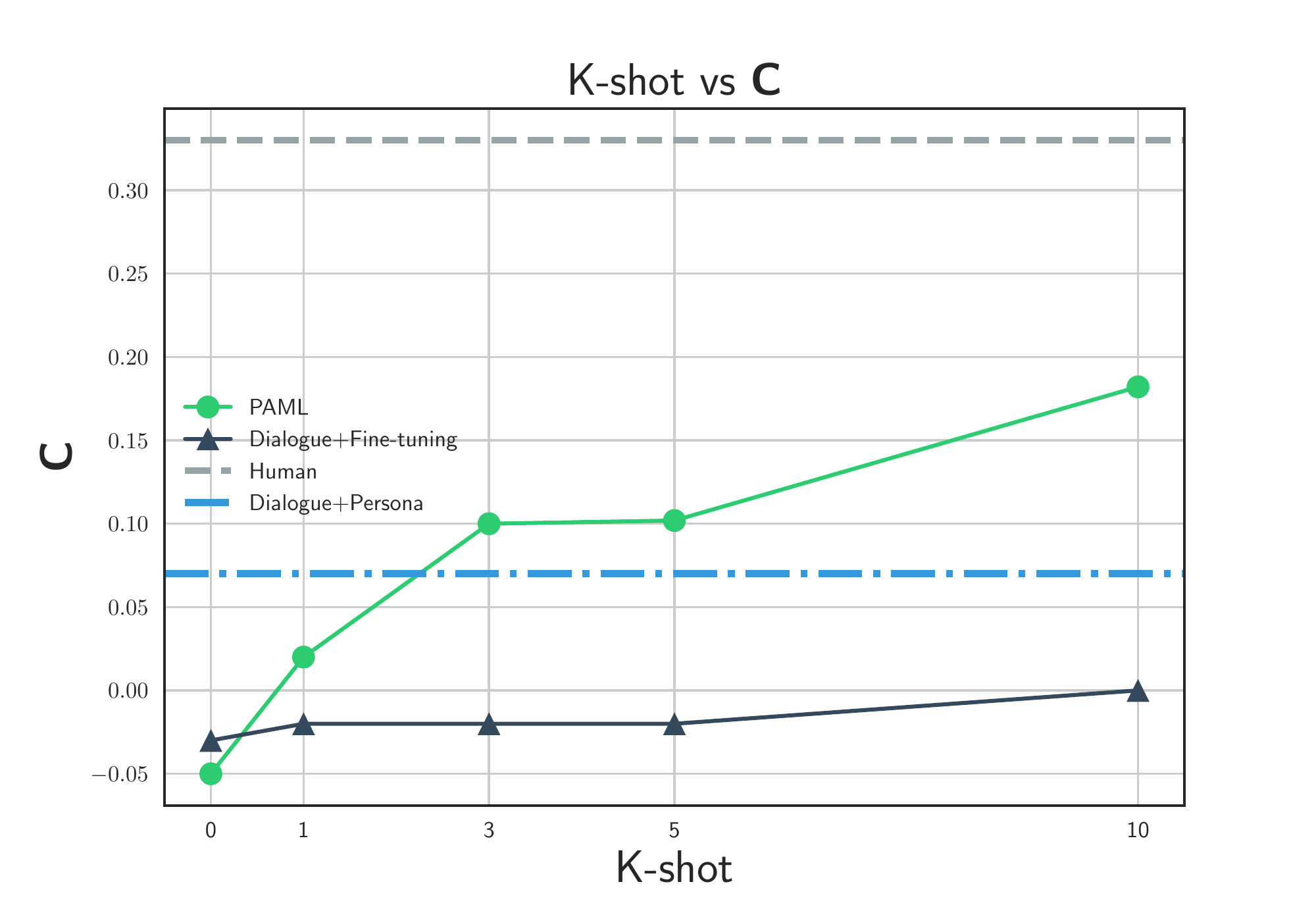}
    \caption{k-shot results for different settings. Consistency of \textit{PAML} grows linearly with respect to $k$.}
    \label{fig:kshot}
\end{figure}

\paragraph{Few-shot Learning}
We analyze the ability of our model to fast adapt to a certain persona in term of shots. We define shot as the number of dialogues used in $\mathcal{D}^{train}_{p_i}$ for fine-tuning a certain persona, e.g. 1-shot one dialogue, 3-shot three dialogue and so on. Figure~\ref{fig:kshot} compares the $k$-shot consistency \textit{\textbf{C}} results for $k$ equal to 0, 1, 3, 5 and 10,  both \textit{PAML} and \textit{Dialogue+Fine-tuning}. \textit{PAML} can achieve a high consistency score just by using 3 dialogues, which is better than \textit{Persona+Dialogue}. On the other hand, \textit{Dialogue+Fine-tuning} cannot properly leverage the dialogues in $\mathcal{D}_{p_i}$, which proves the effectiveness of training with meta-learning. 
\section{Related Work}
\paragraph{Meta-Learning}
Meta-learning~\cite{Thrun:1998:LL:296635,schmidhuber:1987:srl,schmidhuber1992learning,naik1992meta,bengio1992optimization} is sub-field of machine learning with the aim of learning the learning algorithm itself. Recently, several meta-learning models has been proposed for solving few-shot image classification~\cite{ravi2016optimization,vinyals2016matching,finn2017model,mishra2017simple,santoro2016meta}, optimization~\cite{andrychowicz2016learning} and reinforcement learning~\cite{finn2017model}. Meta-learning for NLP application is less common, and it has been applied in semantic parsing task~\cite{N18-2115}, machine translation for low resource language~\cite{D18-1398}, and for text classification \cite{N18-1109}. To the best of our knowledge, this is the first attempt in adapting meta-learning to personalized dialogue learning.

\paragraph{Personalized Dialogue}
\citet{li2016persona} was the first to propose a persona based dialogue models for improving response consistency. \citet{personachat} introduced Persona-chat, which was further extended in ConvAI2~\citeyearpar{dinan2019second}. Several works improved on the initial baselines with various methodologies~\cite{kulikov2018importance,yavuzdeepcopy,hancock2019learning,lucas2009managing,joshi2017personalization, zemlyanskiy2018aiming,gao2018neural}. However, all of these previous works conditioned their response on the persona description, instead of using the dialogues produced by the persona.

\section{Conclusion}
In this paper, we present a novel meta-learning setting for personalizing dialogue agents without conditioning the model response to the persona description. This is especially useful since obtaining such persona description requires human effort. Moreover, we show that a dialogue agent trained with meta-learning achieves a more consistent dialogue by both of automatic measures and human evaluation. In future works, we plan to apply meta-learning to comment generation~\cite{lin2019learning} and task-oriented dialogues systems~\cite{madotto2018mem2seq,wu2018globaltolocal,wu2017end,wu2018end,reddy2018multi}.

\section{Acknowledgments}
This work has been funded by MRP/055/18 of the Innovation Technology Commission, of the Hong Kong University of Science and Technology.

\bibliography{acl2019}

\begin{thebibliography}{39}
\expandafter\ifx\csname natexlab\endcsname\relax\def\natexlab#1{#1}\fi

\bibitem[{Andrychowicz et~al.(2016)Andrychowicz, Denil, Gomez, Hoffman, Pfau,
  Schaul, Shillingford, and De~Freitas}]{andrychowicz2016learning}
Marcin Andrychowicz, Misha Denil, Sergio Gomez, Matthew~W Hoffman, David Pfau,
  Tom Schaul, Brendan Shillingford, and Nando De~Freitas. 2016.
\newblock Learning to learn by gradient descent by gradient descent.
\newblock In \emph{Advances in Neural Information Processing Systems}, pages
  3981--3989.

\bibitem[{Bengio et~al.(1992)Bengio, Bengio, Cloutier, and
  Gecsei}]{bengio1992optimization}
Samy Bengio, Yoshua Bengio, Jocelyn Cloutier, and Jan Gecsei. 1992.
\newblock On the optimization of a synaptic learning rule.
\newblock In \emph{Preprints Conf. Optimality in Artificial and Biological
  Neural Networks}, pages 6--8. Univ. of Texas.

\bibitem[{Chen et~al.(2017)Chen, Zhu, Ling, Wei, Jiang, and Inkpen}]{P17-1152}
Qian Chen, Xiaodan Zhu, Zhen-Hua Ling, Si~Wei, Hui Jiang, and Diana Inkpen.
  2017.
\newblock \href {https://doi.org/10.18653/v1/P17-1152} {Enhanced lstm for
  natural language inference}.
\newblock In \emph{Proceedings of the 55th Annual Meeting of the Association
  for Computational Linguistics (Volume 1: Long Papers)}, pages 1657--1668.
  Association for Computational Linguistics.

\bibitem[{Devlin et~al.(2018)Devlin, Chang, Lee, and
  Toutanova}]{devlin2018bert}
Jacob Devlin, Ming-Wei Chang, Kenton Lee, and Kristina Toutanova. 2018.
\newblock Bert: Pre-training of deep bidirectional transformers for language
  understanding.
\newblock \emph{arXiv preprint arXiv:1810.04805}.

\bibitem[{Dinan et~al.(2019)Dinan, Logacheva, Malykh, Miller, Shuster, Urbanek,
  Kiela, Szlam, Serban, Lowe et~al.}]{dinan2019second}
Emily Dinan, Varvara Logacheva, Valentin Malykh, Alexander Miller, Kurt
  Shuster, Jack Urbanek, Douwe Kiela, Arthur Szlam, Iulian Serban, Ryan Lowe,
  et~al. 2019.
\newblock The second conversational intelligence challenge (convai2).
\newblock \emph{arXiv preprint arXiv:1902.00098}.

\bibitem[{Finn et~al.(2017)Finn, Abbeel, and Levine}]{finn2017model}
Chelsea Finn, Pieter Abbeel, and Sergey Levine. 2017.
\newblock Model-agnostic meta-learning for fast adaptation of deep networks.
\newblock In \emph{Proceedings of the 34th International Conference on Machine
  Learning-Volume 70}, pages 1126--1135. JMLR. org.

\bibitem[{Gao et~al.(2018)Gao, Galley, and Li}]{gao2018neural}
Jianfeng Gao, Michel Galley, and Lihong Li. 2018.
\newblock Neural approaches to conversational ai.
\newblock In \emph{The 41st International ACM SIGIR Conference on Research \&
  Development in Information Retrieval}, pages 1371--1374. ACM.

\bibitem[{Gu et~al.(2018)Gu, Wang, Chen, Li, and Cho}]{D18-1398}
Jiatao Gu, Yong Wang, Yun Chen, Victor O.~K. Li, and Kyunghyun Cho. 2018.
\newblock \href {http://aclweb.org/anthology/D18-1398} {Meta-learning for
  low-resource neural machine translation}.
\newblock In \emph{Proceedings of the 2018 Conference on Empirical Methods in
  Natural Language Processing}, pages 3622--3631. Association for Computational
  Linguistics.

\bibitem[{Hancock et~al.(2019)Hancock, Bordes, Mazare, and
  Weston}]{hancock2019learning}
Braden Hancock, Antoine Bordes, Pierre-Emmanuel Mazare, and Jason Weston. 2019.
\newblock Learning from dialogue after deployment: Feed yourself, chatbot!
\newblock \emph{arXiv preprint arXiv:1901.05415}.

\bibitem[{Huang et~al.(2018)Huang, Wang, Singh, Yih, and He}]{N18-2115}
Po-Sen Huang, Chenglong Wang, Rishabh Singh, Wen-tau Yih, and Xiaodong He.
  2018.
\newblock \href {https://doi.org/10.18653/v1/N18-2115} {Natural language to
  structured query generation via meta-learning}.
\newblock In \emph{Proceedings of the 2018 Conference of the North American
  Chapter of the Association for Computational Linguistics: Human Language
  Technologies, Volume 2 (Short Papers)}, pages 732--738. Association for
  Computational Linguistics.

\bibitem[{Joshi et~al.(2017)Joshi, Mi, and Faltings}]{joshi2017personalization}
Chaitanya~K Joshi, Fei Mi, and Boi Faltings. 2017.
\newblock Personalization in goal-oriented dialog.
\newblock \emph{arXiv preprint arXiv:1706.07503}.

\bibitem[{Kingma and Ba(2014)}]{kingma2014adam}
Diederik~P Kingma and Jimmy Ba. 2014.
\newblock Adam: A method for stochastic optimization.
\newblock \emph{arXiv preprint arXiv:1412.6980}.

\bibitem[{Kulikov et~al.(2018)Kulikov, Miller, Cho, and
  Weston}]{kulikov2018importance}
Ilya Kulikov, Alexander~H Miller, Kyunghyun Cho, and Jason Weston. 2018.
\newblock Importance of a search strategy in neural dialogue modelling.
\newblock \emph{arXiv preprint arXiv:1811.00907}.

\bibitem[{Li et~al.(2016)Li, Galley, Brockett, Spithourakis, Gao, and
  Dolan}]{li2016persona}
Jiwei Li, Michel Galley, Chris Brockett, Georgios Spithourakis, Jianfeng Gao,
  and Bill Dolan. 2016.
\newblock A persona-based neural conversation model.
\newblock In \emph{Proceedings of the 54th Annual Meeting of the Association
  for Computational Linguistics (Volume 1: Long Papers)}, volume~1, pages
  994--1003.

\bibitem[{Lin et~al.(2019)Lin, Winata, and Fung}]{lin2019learning}
Zhaojiang Lin, Genta~Indra Winata, and Pascale Fung. 2019.
\newblock Learning comment generation by leveraging user-generated data.
\newblock In \emph{ICASSP 2019-2019 IEEE International Conference on Acoustics,
  Speech and Signal Processing (ICASSP)}, pages 7225--7229. IEEE.

\bibitem[{Liu et~al.(2016)Liu, Lowe, Serban, Noseworthy, Charlin, and
  Pineau}]{D16-1230}
Chia-Wei Liu, Ryan Lowe, Iulian Serban, Mike Noseworthy, Laurent Charlin, and
  Joelle Pineau. 2016.
\newblock \href {https://doi.org/10.18653/v1/D16-1230} {How not to evaluate
  your dialogue system: An empirical study of unsupervised evaluation metrics
  for dialogue response generation}.
\newblock In \emph{Proceedings of the 2016 Conference on Empirical Methods in
  Natural Language Processing}, pages 2122--2132. Association for Computational
  Linguistics.

\bibitem[{Lucas et~al.(2009)Lucas, Fern{\'a}ndez, Salazar, Ferreiros, and
  San~Segundo}]{lucas2009managing}
JM~Lucas, F~Fern{\'a}ndez, J~Salazar, J~Ferreiros, and R~San~Segundo. 2009.
\newblock Managing speaker identity and user profiles in a spoken dialogue
  system.
\newblock \emph{Procesamiento del lenguaje natural}, (43).

\bibitem[{Madotto et~al.(2018)Madotto, Wu, and Fung}]{madotto2018mem2seq}
Andrea Madotto, Chien-Sheng Wu, and Pascale Fung. 2018.
\newblock Mem2seq: Effectively incorporating knowledge bases into end-to-end
  task-oriented dialog systems.
\newblock In \emph{Proceedings of the 56th Annual Meeting of the Association
  for Computational Linguistics (Volume 1: Long Papers)}, pages 1468--1478.

\bibitem[{Mazare et~al.(2018)Mazare, Humeau, Raison, and
  Bordes}]{millionspersona}
Pierre-Emmanuel Mazare, Samuel Humeau, Martin Raison, and Antoine Bordes. 2018.
\newblock \href {http://aclweb.org/anthology/D18-1298} {Training millions of
  personalized dialogue agents}.
\newblock In \emph{Proceedings of the 2018 Conference on Empirical Methods in
  Natural Language Processing}, pages 2775--2779. Association for Computational
  Linguistics.

\bibitem[{Mishra et~al.(2017)Mishra, Rohaninejad, Chen, and
  Abbeel}]{mishra2017simple}
Nikhil Mishra, Mostafa Rohaninejad, Xi~Chen, and Pieter Abbeel. 2017.
\newblock A simple neural attentive meta-learner.
\newblock \emph{ICLR}.

\bibitem[{Naik and Mammone(1992)}]{naik1992meta}
Devang~K Naik and RJ~Mammone. 1992.
\newblock Meta-neural networks that learn by learning.
\newblock In \emph{[Proceedings 1992] IJCNN International Joint Conference on
  Neural Networks}, volume~1, pages 437--442. IEEE.

\bibitem[{Papineni et~al.(2002)Papineni, Roukos, Ward, and
  Zhu}]{papineni2002bleu}
Kishore Papineni, Salim Roukos, Todd Ward, and Wei-Jing Zhu. 2002.
\newblock Bleu: a method for automatic evaluation of machine translation.
\newblock In \emph{Proceedings of the 40th annual meeting on association for
  computational linguistics}, pages 311--318. Association for Computational
  Linguistics.

\bibitem[{Pennington et~al.(2014)Pennington, Socher, and
  Manning}]{pennington2014glove}
Jeffrey Pennington, Richard Socher, and Christopher Manning. 2014.
\newblock Glove: Global vectors for word representation.
\newblock In \emph{Proceedings of the 2014 conference on empirical methods in
  natural language processing (EMNLP)}, pages 1532--1543.

\bibitem[{Ravi and Larochelle(2016)}]{ravi2016optimization}
Sachin Ravi and Hugo Larochelle. 2016.
\newblock Optimization as a model for few-shot learning.

\bibitem[{Reddy et~al.(2018)Reddy, Contractor, Raghu, and
  Joshi}]{reddy2018multi}
Revanth Reddy, Danish Contractor, Dinesh Raghu, and Sachindra Joshi. 2018.
\newblock Multi-level memory for task oriented dialogs.
\newblock \emph{arXiv preprint arXiv:1810.10647}.

\bibitem[{Santoro et~al.(2016)Santoro, Bartunov, Botvinick, Wierstra, and
  Lillicrap}]{santoro2016meta}
Adam Santoro, Sergey Bartunov, Matthew Botvinick, Daan Wierstra, and Timothy
  Lillicrap. 2016.
\newblock Meta-learning with memory-augmented neural networks.
\newblock In \emph{International conference on machine learning}, pages
  1842--1850.

\bibitem[{Schmidhuber(1987)}]{schmidhuber:1987:srl}
Jurgen Schmidhuber. 1987.
\newblock \href {http://www.idsia.ch/~juergen/diploma.html} {Evolutionary
  principles in self-referential learning. on learning now to learn: The
  meta-meta-meta...-hook}.
\newblock Diploma thesis, Technische Universitat Munchen, Germany, 14 May.

\bibitem[{Schmidhuber(1992)}]{schmidhuber1992learning}
J{\"u}rgen Schmidhuber. 1992.
\newblock Learning to control fast-weight memories: An alternative to dynamic
  recurrent networks.
\newblock \emph{Neural Computation}, 4(1):131--139.

\bibitem[{Sean et~al.(2018)Sean, Weston, Szlam, and Cho}]{dnli}
Welleck Sean, Jason Weston, Arthur Szlam, and Kyunghyun Cho. 2018.
\newblock Dialogue natural language inference.
\newblock \emph{arXiv preprint arXiv:1811.00671}.

\bibitem[{Thrun and Pratt(1998)}]{Thrun:1998:LL:296635}
Sebastian Thrun and Lorien Pratt, editors. 1998.
\newblock \emph{Learning to Learn}.
\newblock Kluwer Academic Publishers, Norwell, MA, USA.

\bibitem[{Vaswani et~al.(2017)Vaswani, Shazeer, Parmar, Uszkoreit, Jones,
  Gomez, Kaiser, and Polosukhin}]{transformer}
Ashish Vaswani, Noam Shazeer, Niki Parmar, Jakob Uszkoreit, Llion Jones,
  Aidan~N Gomez, \L~ukasz Kaiser, and Illia Polosukhin. 2017.
\newblock \href
  {http://papers.nips.cc/paper/7181-attention-is-all-you-need.pdf} {Attention
  is all you need}.
\newblock In I.~Guyon, U.~V. Luxburg, S.~Bengio, H.~Wallach, R.~Fergus,
  S.~Vishwanathan, and R.~Garnett, editors, \emph{Advances in Neural
  Information Processing Systems 30}, pages 5998--6008. Curran Associates, Inc.

\bibitem[{Vinyals et~al.(2016)Vinyals, Blundell, Lillicrap, Wierstra
  et~al.}]{vinyals2016matching}
Oriol Vinyals, Charles Blundell, Timothy Lillicrap, Daan Wierstra, et~al. 2016.
\newblock Matching networks for one shot learning.
\newblock In \emph{Advances in neural information processing systems}, pages
  3630--3638.

\bibitem[{Wu et~al.(2017)Wu, Madotto, Winata, and Fung}]{wu2017end}
Chien-Sheng Wu, Andrea Madotto, Genta Winata, and Pascale Fung. 2017.
\newblock End-to-end recurrent entity network for entity-value independent
  goal-oriented dialog learning.
\newblock In \emph{Dialog System Technology Challenges Workshop, DSTC6}.

\bibitem[{Wu et~al.(2018)Wu, Madotto, Winata, and Fung}]{wu2018end}
Chien-Sheng Wu, Andrea Madotto, Genta~Indra Winata, and Pascale Fung. 2018.
\newblock End-to-end dynamic query memory network for entity-value independent
  task-oriented dialog.
\newblock In \emph{2018 IEEE International Conference on Acoustics, Speech and
  Signal Processing (ICASSP)}, pages 6154--6158. IEEE.

\bibitem[{Wu et~al.(2019)Wu, Socher, and Xiong}]{wu2018globaltolocal}
Chien-Sheng Wu, Richard Socher, and Caiming Xiong. 2019.
\newblock \href {https://openreview.net/forum?id=ryxnHhRqFm} {Global-to-local
  memory pointer networks for task-oriented dialogue}.
\newblock In \emph{International Conference on Learning Representations}.

\bibitem[{Yavuz et~al.()Yavuz, Rastogi, Chao, Hakkani-T{\"u}r, and
  AI}]{yavuzdeepcopy}
Semih Yavuz, Abhinav Rastogi, Guan-lin Chao, Dilek Hakkani-T{\"u}r, and
  Amazon~Alexa AI.
\newblock Deepcopy: Grounded response generation with hierarchical pointer
  networks.

\bibitem[{Yu et~al.(2018)Yu, Guo, Yi, Chang, Potdar, Cheng, Tesauro, Wang, and
  Zhou}]{N18-1109}
Mo~Yu, Xiaoxiao Guo, Jinfeng Yi, Shiyu Chang, Saloni Potdar, Yu~Cheng, Gerald
  Tesauro, Haoyu Wang, and Bowen Zhou. 2018.
\newblock \href {https://doi.org/10.18653/v1/N18-1109} {Diverse few-shot text
  classification with multiple metrics}.
\newblock In \emph{Proceedings of the 2018 Conference of the North American
  Chapter of the Association for Computational Linguistics: Human Language
  Technologies, Volume 1 (Long Papers)}, pages 1206--1215. Association for
  Computational Linguistics.

\bibitem[{Zemlyanskiy and Sha(2018)}]{zemlyanskiy2018aiming}
Yury Zemlyanskiy and Fei Sha. 2018.
\newblock Aiming to know you better perhaps makes me a more engaging dialogue
  partner.
\newblock \emph{CoNLL 2018}, page 551.

\bibitem[{Zhang et~al.(2018)Zhang, Dinan, Urbanek, Szlam, Kiela, and
  Weston}]{personachat}
Saizheng Zhang, Emily Dinan, Jack Urbanek, Arthur Szlam, Douwe Kiela, and Jason
  Weston. 2018.
\newblock \href {http://aclweb.org/anthology/P18-1205} {Personalizing dialogue
  agents: I have a dog, do you have pets too?}
\newblock In \emph{Proceedings of the 56th Annual Meeting of the Association
  for Computational Linguistics (Volume 1: Long Papers)}, pages 2204--2213.
  Association for Computational Linguistics.

\end{thebibliography}
\bibliographystyle{acl_natbib}

\appendix

\section{Supplemental Material}
\label{sec:supplemental}

\subsection{Dialogue examples}

\begin{table*}[b]
\begin{tabular}{rl}
\hline
\multicolumn{2}{c}{\textbf{Persona}} \\ \hline
\multicolumn{2}{c}{\begin{tabular}[c]{@{}c@{}}autumn is my favorite season \\ i have a turtle named timothy \\ my favorite sport is ultimate frisbee \\ my parents are living in bora bora\end{tabular}} \\ \hline
\multicolumn{2}{c}{\textbf{Dialogue History}} \\ \hline
\multicolumn{1}{r|}{\textit{Usr:}} & hey there . how are you ? \\
\multicolumn{1}{r|}{\textit{Sys:}} & hey there . i am fine . what about you ? \\
\multicolumn{1}{r|}{\textit{Usr:}} & been better . stressful day at work . \\
\multicolumn{1}{r|}{\textit{Sys:}} & do you work on sundays ? \\
\multicolumn{1}{r|}{\textit{Usr:}} & yep . i work fast food . always working on a sunday . \\
\multicolumn{1}{r|}{\textit{Sys:}} & how old are you ? if do not mind to share it . \\
\multicolumn{1}{r|}{\textit{Usr:}} & i am 20 and i have been doing this for 4 years . \\
\multicolumn{1}{r|}{\textit{Sys:}} & you are too young . do you study ? \\
\multicolumn{1}{r|}{\textit{Usr:}} & \begin{tabular}[c]{@{}l@{}}no . i left high school at 16 to help my family . trying \\ to get my ged soon .\end{tabular} \\
\multicolumn{1}{r|}{\textit{Sys:}} & i feel so sorry . you should get it soon . \\
\multicolumn{1}{r|}{\textit{Usr:}} & \begin{tabular}[c]{@{}l@{}}yeah trying my best but helping my family out of \\ these bad apartments is my main goal .\end{tabular} \\
\multicolumn{1}{r|}{\textit{Sys:}} & god help you and make you rich one day . \\
\multicolumn{1}{r|}{\textit{Usr:}} & haha thanks . enough about me . what do you like to do ? \\ \hline
\multicolumn{2}{c}{\textbf{Generated Responses}} \\ \hline
\multicolumn{1}{r|}{\textit{Human:}} & i finished college now looking for a job \\ \hline
\multicolumn{1}{r|}{\textit{Dialogue:}} & i like to play video games . \\
\multicolumn{1}{r|}{\textit{Persona+Dialogue:}} & i like to play ultimate frisbee . \\
\multicolumn{1}{r|}{\textit{Multi-task+Fine-tune:}} & i like to go to the movies . \\
\multicolumn{1}{r|}{\textit{PAML:}} & i like to play frisbee with my friends . \\ \hline
\end{tabular}
\end{table*}

\begin{table*}[t]
\begin{tabular}{rl}
\hline
\multicolumn{2}{c}{\textbf{Persona}} \\ \hline
\multicolumn{2}{c}{\begin{tabular}[c]{@{}c@{}}i am a student \\ i am very studious and do not like to party or drink \\ i grew up in the united states of america \\ i want to be a marine biologist\end{tabular}} \\ \hline
\multicolumn{2}{c}{\textbf{Dialogue history}} \\ \hline
\multicolumn{1}{r|}{\textit{Usr:}} & hi how are you today \\
\multicolumn{1}{r|}{\textit{Sys:}} & i am well , studying for an exam \\
\multicolumn{1}{r|}{\textit{Usr:}} & for college or for what ? \\
\multicolumn{1}{r|}{\textit{Sys:}} & \begin{tabular}[c]{@{}l@{}}yes college . working on a degree in marine \\ biology\end{tabular} \\
\multicolumn{1}{r|}{\textit{Usr:}} & nice i love fish and that stuff \\
\multicolumn{1}{r|}{\textit{Sys:}} & \begin{tabular}[c]{@{}l@{}}sounds fun , unlike kids my age i do not \\ drink , smoke or party\end{tabular} \\
\multicolumn{1}{r|}{\textit{Usr:}} & straight edge is the way to go minor threat says \\
\multicolumn{1}{r|}{\textit{Sys:}} & i guess . what do you do \\
\multicolumn{1}{r|}{\textit{Usr:}} & i just go to the gym all day really \\
\multicolumn{1}{r|}{\textit{Sys:}} & you must be very strong then \\
\multicolumn{1}{r|}{\textit{Usr:}} & yeah but i am single . \\
\multicolumn{1}{r|}{\textit{Sys:}} & trying to find a mate at the gym ? \\
\multicolumn{1}{r|}{\textit{Usr:}} & no i actually enjoy being alone \\ \hline
\multicolumn{2}{c}{\textbf{Generated Responses}} \\ \hline
\multicolumn{1}{r|}{\textit{Human:}} & \begin{tabular}[c]{@{}l@{}}that is absolutely fine nothing wrong \\ with being alone\end{tabular} \\
\multicolumn{1}{r|}{\textit{Dialog:}} & what do you do for work ? \\
\multicolumn{1}{r|}{\textit{Persona+Dialog:}} & what do you do for work ? \\
\multicolumn{1}{r|}{\textit{Multi-task+Finetune:}} & what do you do for work ? \\
\multicolumn{1}{r|}{\textit{PAML:}} & do you like to read ? i like to read . \\ \hline
\end{tabular}
\end{table*}

\begin{table*}[t]
\begin{tabular}{rl}
\hline
\multicolumn{2}{c}{\textbf{Persona}} \\ \hline
\multicolumn{2}{c}{\begin{tabular}[c]{@{}c@{}}autumn is my favorite season \\ i have a turtle named timothy \\ i love to meet new people \\ my parents are living in bora bora\end{tabular}} \\ \hline
\multicolumn{2}{c}{\textbf{Dialogue history}} \\ \hline
\multicolumn{1}{r|}{\textit{Usr:}} & hi , tell me about yourself . \\
\multicolumn{1}{r|}{\textit{Sys:}} & i have a pet turtle that i love ! his name is timothy \\
\multicolumn{1}{r|}{\textit{Usr:}} & that is cool . i like fantasy videogames . do you play any ? \\
\multicolumn{1}{r|}{\textit{Sys:}} & not really . i like hanging out with people in person \\
\multicolumn{1}{r|}{\textit{Usr:}} & \begin{tabular}[c]{@{}l@{}}ah . well i am pretty shy so i do not go out with \\ people much .\end{tabular} \\
\multicolumn{1}{r|}{\textit{Sys:}} & i understand . i love autumn , how about you ? \\
\multicolumn{1}{r|}{\textit{Usr:}} & \begin{tabular}[c]{@{}l@{}}yeah , i am from the south so its nice when it cools off \\ for a while .\end{tabular} \\
\multicolumn{1}{r|}{\textit{Sys:}} & definitely . my parents went away to bora bora for a while . \\
\multicolumn{1}{r|}{\textit{Usr:}} & oh cool ! my mom is a doctor so my parents travel a lot . \\
\multicolumn{1}{r|}{\textit{Sys:}} & what do you do for a living ? \\
\multicolumn{1}{r|}{\textit{Usr:}} & \begin{tabular}[c]{@{}l@{}}i am a computer engineer got a comp sci degree . \\ what do you do ?\end{tabular} \\
\multicolumn{1}{r|}{\textit{Sys:}} & i am currently in school \\
\multicolumn{1}{r|}{\textit{Usr:}} & do you have any hobbies ? i build model spaceships . \\ \hline
\multicolumn{2}{c}{\textbf{Generated Responses}} \\ \hline
\multicolumn{1}{r|}{\textit{Human:}} & i guess i like to cook \\
\multicolumn{1}{r|}{\textit{Dialog:}} & that sounds like a lot of fun . \\
\multicolumn{1}{r|}{\textit{Persona+Dialog:}} & i love to build people . \\
\multicolumn{1}{r|}{\textit{Multi-task+Finetune:}} & that is a good idea . i love to play the guitar . \\
\multicolumn{1}{r|}{\textit{PAML:}} & that is awesome . i like to play video games . \\ \hline
\end{tabular}
\end{table*}

\subsection{Plots}
\begin{figure}[H]
  \centering
  \includegraphics[width=1.0\linewidth]{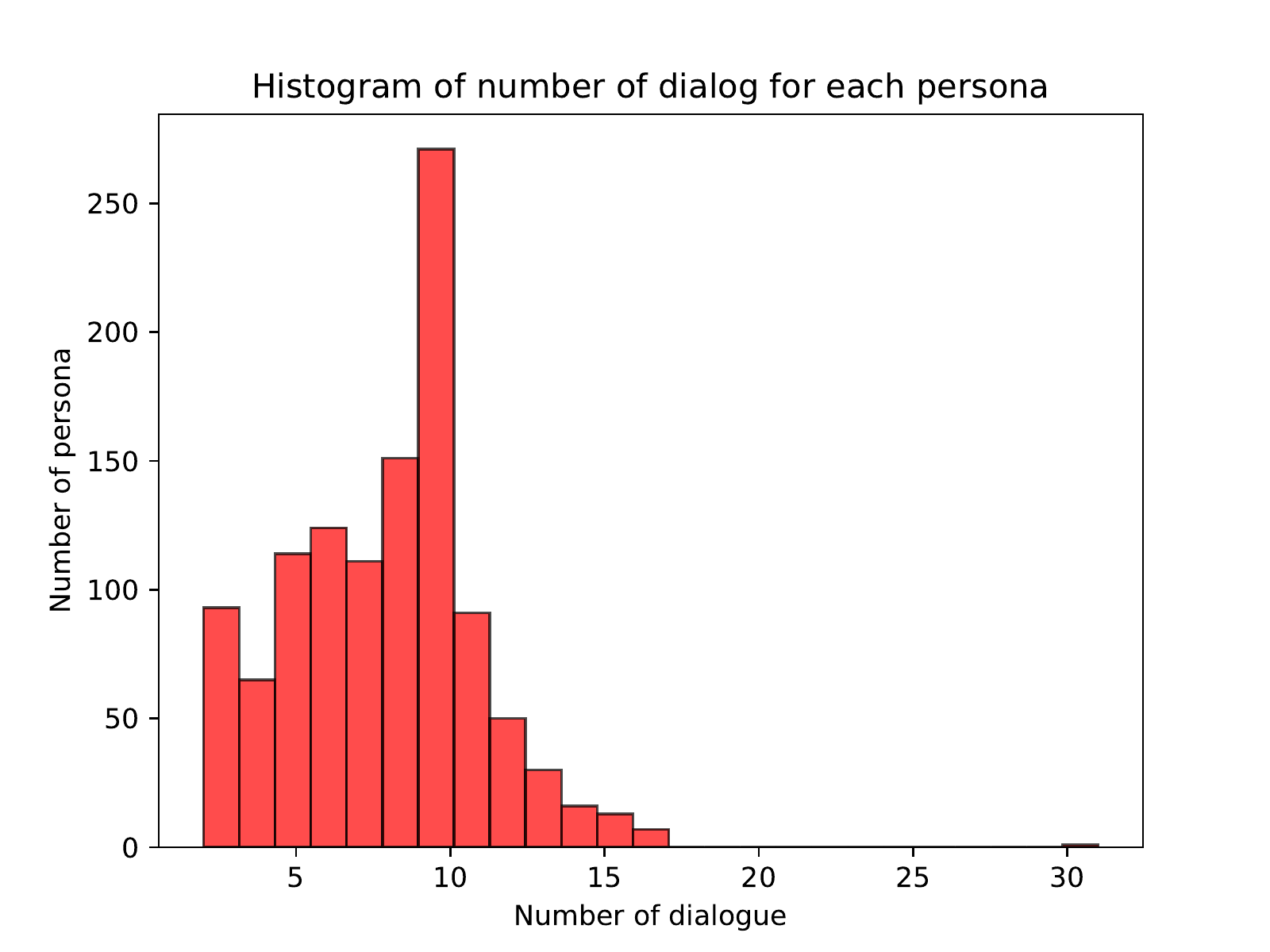}
  \caption{Distribution of number of dialogues for each persona description in the training set.}
  \label{fig:hist}
\end{figure}

\begin{figure}[H]
    \centering
    \includegraphics[width=\linewidth]{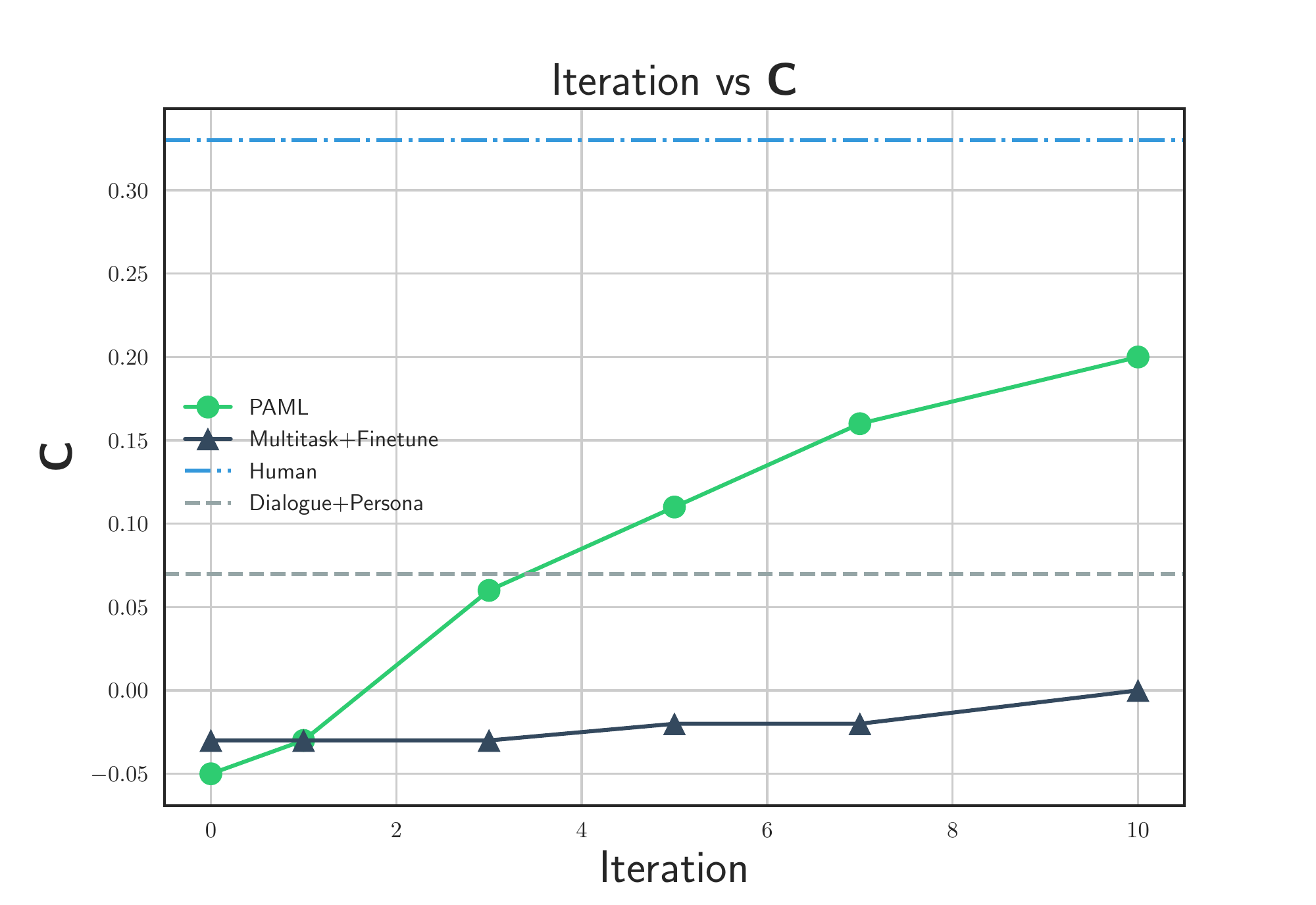}
    \caption{Iteration of finetuning versus consistency. Consistency of \textit{PAML} grows linearly with respect to the iteration.}
    \label{fig:my_label}
\end{figure}

\subsection{Human evaluation}
To each crowed worker we show a dialogue history, a persona description and the generated response from one of the evaluated settings. Then we ask them to evaluate fluency and consistency. The former is a pretty straightforward measure, where instead we defined consistency as following:

An answer is considered \textbf{consistent} if and only if it 
\begin{itemize}
    \item does \textbf{not contradict} with neither the dialogue history, nor the persona description;
    \item is \textbf{relevant} to any of the given persona description sentences.
\end{itemize}
Usually, generic answer like "I am not sure" or "I am sorry to hear that" are considered Neutral. For example, from the persona description, if User 2 likes basketball, talking about basketball will make the answer \textbf{consistent}. An answer like "I hate basketball" will be considered a \textbf{contradiction}.
However, in the following cases, the answer is considered \textbf{neutral}:
\begin{itemize}
    \item The answer does \textbf{not contradict} neither the dialogue history nor the persona description
    \item The answer is \textbf{not relevant} to any of the given persona description sentences
\end{itemize}
For example, from the persona description, if User 2 likes basketball, talking about swimming is considered \textbf{neutral}, as it is not relevant to basketball but does not contradict anything.

Therefore, we ask you to score only the consistency as such:
\begin{itemize}
    \item The answer is contradicting:-1 
    \item The answer is neutral: 0 
    \item The answer is consistent: 1 
\end{itemize}

\end{document}